# A Soft High Force Hand Exoskeleton for Rehabilitation and Assistance of Spinal Cord Injury and Stroke Individuals

Shuangyue Yu, Hadia Perez, James Barkas, Mohamed Mohamed, Mohamed Eldaly,
Tzu-Hao Huang, Xiaolong Yang, Hao Su*
Lab of Biomechatronic and Intelligent Robotics
Grove School of Engineering
City College of New York
New York, NY, USA

Maria del Mar Cortes
Icahn School of Medicine at Mount Sinai
New York, NY, USA

Dylan J. Edwards
Moss Rehabilitation Research Institute
Elkins Park, PA, USA

## ABSTRACT

*Individuals with spinal cord injury (SCI) and stroke who is lack of manipulation capability have a particular need for robotic hand exoskeletons. Among assistive and rehabilitative medical exoskeletons, there exists a sharp trade-off between device power on the one hand and ergonomics and portability on other, devices that provide stronger grasping assistance do so at the cost of patient comfort. This paper proposes using fin-ray-inspired, cable-driven finger orthoses to generate high fingertip forces without the painful compressive and shear stresses commonly associated with conventional cable-drive exoskeletons. With combination cable-driven transmission and segmented-finger orthoses, the exoskeleton transmitted larger forces and applied torques discretely to the fingers, leading to strong fingertip forces. A prototype of the finger orthoses and associated cable transmission was fabricated, and force transmission tests of the prototype in the finger flexion mode demonstrated a 2:1 input-output ratio between cable tension and fingertip force, with a maximum fingertip force of 22 N. Moreover, the proposed design provides a comfortable experience for wearers thanks to its lightweight and conformal properties to the hands.*

## INTRODUCTION

In the U.S. approximately 282,000 people have SCI with 17,000 new cases each year [1]. SCI that occur above the C8 vertebrae may cause severe losses of hand strength and dexterity [2]. As a result, patients will experience different difficulties with activities of daily living (ADL). Individuals with SCI might not be able to complete manipulation tasks. Existing soft exoskeletons provide limited grip and pinch strength, with the highest-performing devices achieving around 10 N for each finger, less than is required for several ADLs. This lack of strength is not sufficient to generate a firm grip on objects [3-5]. The design objective is to provide assistance to individuals with SCI and stroke who lose hand function by restoring flexion and extension to the fingers. This has the potential to provide long-lasting use of the device through a portable actuation unit that can be mounted on wheelchairs or other mobility equipment.

## METHODS

Functional requirements were formulated for the hand exoskeleton, as seen in Table 1. First, the device should be able to actively support flexion and extension of the fingers. The range of motion (ROM) is different for the fingers and the thumb, as seen below in Table 1. Furthermore, the device should provide passive support for adduction/abduction. Additionally, the device should be able to optimally transfer force from the actuation unit to the finger without compressing the finger or otherwise causing pain to the patient. For daily activity tasks that involve effective grip, the device should be able to provide about 25 N of fingertip force. This value was derived from a force balance calculation for the case of an individual holding up a 1kg object against gravity using only by the fingertips, using empirically-derived values for static friction coefficients of the fingertip [6], and incorporating a safety factor of 2. Moreover, the device should feel comfortable and safe for patients. To this end, the device should be lightweight (<500g) and should require minimal skin coverage on the hand. Finally, the device should not restrict the patient's mobility and preferably it should be portable. The actuation system should be contained within the wheelchair of the individual.

**Table 1:** Functional requirements of the hand exoskeleton.

| Motion | 1. Passive adduction/abduction<br>2. Active flexion/extension:<br><br>Finger Ranges of Motion [7]:<br><br>|         | Extension   | Flexion     |
|        |---------|-------------|-------------|
|        | MCP all | −19 (SD 6.9)| 90 (SD 9.1) |
|        | PIP all | − 7 (SD 3.7)| 101 (SD 8.3)|
|        | DIP all | − 6 (SD 4.1)| 84 (SD 8.5) |

| | Thumb Range of Motion [8]: |
|---|---|
| | Thumb Joints    Extension    Flexion <br> IP    -12 (SD 9.2)    88 (SD 9.2) <br> MPC    -8.1 (SD 4.4)    60 (SD 5.5) |
| Force Transmission | ○ 25 N of Fingertip Force |
| Comfort Portability | ○ Lightweight (< 500 g) <br> ○ Minimal skin coverage <br> ○ Actuation unit mountable on a wheelchair |

Based on the functional requirements, a hand exoskeleton was designed and assembled, as shown in Fig. 1, which exploits the lightweight, small size, and effective force transmission of tendon-driven soft exoskeletons while achieving higher force transmission than soft exoskeletons. The hand exoskeleton consists of an actuation package, a wearable structure, and a cable transmission structure. The soft robotic glove is safe and lightweight because of soft material structure and innovative high-torque density actuator. The 300g glove has a low profile (154×40×14mm) and lightweight soft structure. Thanks to the lightweight and modular wearable structure design, the subject can easily don/doff the hand exoskeleton.

The ROM of distal interphalangeal (DIP) joints, proximal interphalangeal (PIP) joints, and metacarpophalangeal (MCP) joints of the robotic glove are very similar to the human anatomy, thus robot does not constrain users wearing the soft robotic glove to perform both power and precision grasps, ask shown in Fig. 1.

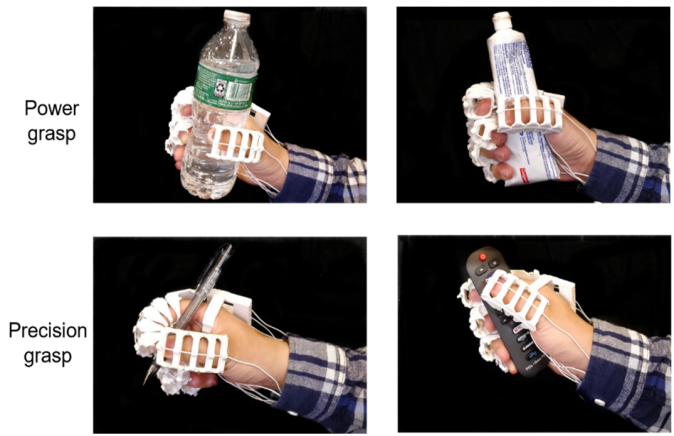

**Fig. 1:** Power and precision grasp demonstration with the soft robotic glove assistance.

The segmented-finger orthosis design used in the exoskeleton is based partially by the work of Kim et. al. [9] on fin-ray-inspired exoskeletons. In contrast to conventional cable-driven soft exoskeletons, in which the cable applies a retractive force to the fingertip only, cable tension in the segmented-orthosis design has the effect of rotating each individual segment in the orthosis. Instead of applying a concentrated bending load to the fingertip, discrete moments are applied along the length of the finger. This series of moments accomplishes three important tasks: 1) the finger is not compressed, which is more comfortable for the patient, 2) the moment on each segment allows the finger to curve more naturally, and 3) applied tension is converted efficiently into a usable moment at the fingertip, transmitting larger forces to the fingers. The phalange segments were made from polylactic acid (PLA) plastic and printed using a MakerBot Replicator FDM 3D printer. Individual segments were printed separately and then assembled. The whole wearable structure doesn't include any electronics components and has very good waterproof ability to be washed repeatedly. A standard 1.28 mm 14-gauge AWG wire with rubber insulation outer layer was used to assemble the segments into phalanges. These wires were run through the two cable passages on the sides of the segments and tied at the ends, creating connections similar in operation to pinned joints.

A wearable cable guide structure was designed to route the cables along the palm of the user in a comfortable and practical manner. This structure was designed to conform to the geometry of a human hand so that it fits comfortably and routes cables along the flexor tendons. The cable guide also serves to minimize shear stress on the patient's palm. The cable guide was made from a photopolymer resin using a stereolithography 3D printer (Formlabs Inc., Form 2). The bases for straps are made to fasten the cable guide along the wrist and along the lower part of the palm for sturdiness and a close fit to the palm.

In the current design, the tension of flexor and extensor cables is powered by an array of electric motor-driven pulleys. The prototype actuation unit consists of four custom-made electric motors, each of which supports an individual flexor cable. The ultimate goal is to design a transmission in which one motor can simultaneously support both a flexor and extensor cable, enabling active assistance for finger flexion and extension. The current prototype incorporates two rerouting pulleys per motor in anticipation of supporting flexor and extensor cables simultaneously.

The actuation pack is lightweight due to the embedded high-torque density actuator [10, 11] can produce the same amount of torque with less than half weight of conventional actuators. This lightweight but high-torque density actuator has been successfully used for lightweight knee exoskeletons [12] and robotic prosthetic hands [11]. This cable-driven actuator possesses inherently interaction-safe feature due to low mechanical impedance and high-sensitivity to the interaction force between the hand and the glove. In contrast to conventional transmission design that typically uses large gear ratio reduction while exhibiting high impedance feature (resistive to user motion), high torque density actuator uses small gear ratio reduction and produces low impedance. Thus, the onboard sensor can proprioceptively monitor interaction between robot and wearer to ensure safety and comfort.

## RESULTS

First, the ratio of cable tension input to fingertip force output during flexion of an individual finger was measured. A torque spring scale was used to measure the tension on the cable as the finger was flexed, while a gravimetric scale was used to measure the fingertip force. The subject wears the hand exoskeleton and the actuation package is placed on the table. The actuator generates certain torque by current control mode, and the torque spring scale and gravimetric scale the cable tension and fingertip force. Fig. 2 shows the results of the experiment. The input force to output force ratio of the device is approximately 2:1. A maximum fingertip force of about 22 N was reached at an input cable tension of 44 N. This shows that this product is cable of producing large output force.

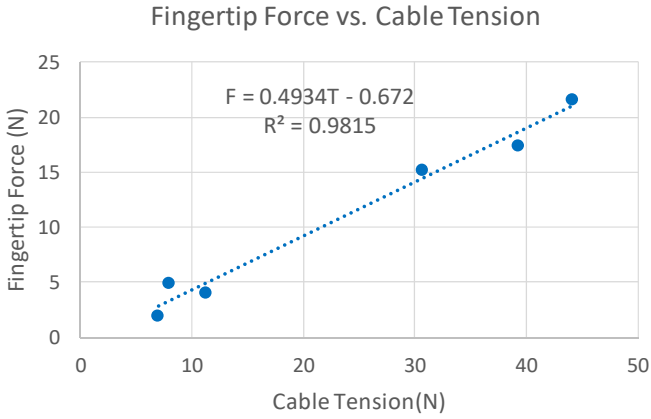

**Fig. 2:** Graphed results of force transmission in finger flexion results and the linear relationship between tension and fingertip force in a 2:1 ratio.

Next, the segmented-phalange structure was evaluated for comparison with experimental results, and to optimize the structure for grip strength, where the fingertip force generated by a segment is the parameter to optimize. At present, only the static flexion case is of interest, however, this analysis can be adapted to the dynamic case.

For the purpose of this analysis, phalange segments are treated as diamond-shaped polygons, as shown in Fig. 3, and the phalange is treated as a planar mechanism consisting of segments linked by frictionless pin joints. In principle, the analysis is applicable to a structure with *n* segments, though for simplicity only a phalange with 3 segments is presented here. The structure is grasping a cylindrical object with a radius of curvature $R$, and the flexor cable is assumed taut, with input tension $T_o$. The structure is fixed at the left-most pin joint $O$, representing the knuckle, and is loaded by a fingertip force $P_{tip}$, by contact forces $P_i$ between the structure and the object at each phalange, by contact forces $N_i$ between the segments and the finger, by cable-induced force $F_i$ in the cable passages of each segment, and by friction forces $f_i$ at each cable passage from the cable.

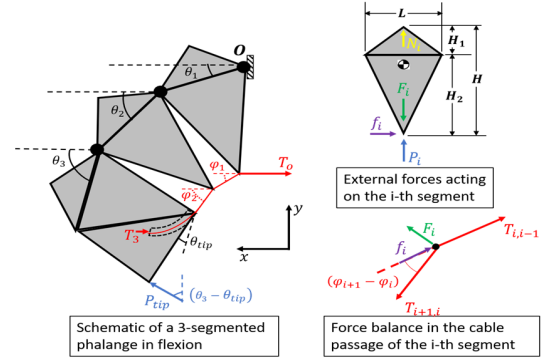

**Fig. 3:** Static analysis of an individual finger during exoskeleton-assisted flexion.

In brief, friction forces generated by the cable can be functionally related to the input tension, and for this case a linear relationship between cable tension $T_i$, friction $f_i$, and the input tension $T_o$ will be assumed, where $c$ is a loss coefficient, assumed constant:

$$T_{i,i+1} = cT_{i-1,i} \quad 0 < c \leq 1 \tag{1}$$

$$T_n = c^n T_o \tag{2}$$

$$f_i = T_{i-1,i} - T_{i,i+1} = c^{(i-1)}(1-c)T_o \tag{3}$$

Similarly, forces $F_i$ from the cable can be related to cable tension through trigonometry and a force balance about the cable passage in any given segment, as shown in Fig. 3:

$$F_i = 2cT_{i,i-1} \sin\left(\frac{1}{2}[\varphi_{i+1} - \varphi_i]\right) \tag{4}$$

For a given object with a radius of curvature $R$, relationships between the angles of rotation of the segments and the flexor cable may be inferred if it is assumed that the phalange conforms perfectly to the curvature of the object:

$$\theta_i = \frac{nL}{R}\left(1 - \frac{i}{n+1}\right) \tag{5}$$

$$\sin\varphi_i = \frac{\cos(\delta - \theta_{i+1}) - \cos\delta}{2\sin(\delta - \theta_{i+1}/2)} \tag{6}$$

where $\delta = \tan^{-1}(L/2H_2)$.

For the static case, and assuming only the fingertip is in contact with the grasped object, contact forces between the segments, the finger and the grasped object may be ignored ($P_i = N_i = 0$). Taking the moment balance about $O$, the fingertip force $P_{tip}$ can be obtained from the moments generated by the cable tension, the cable pressure, and friction forces:

$$\left|\vec{P}_{tip} \times \vec{r}_{tip}\right| = \left|M_{T_n}\right| + \left|M_{F_i}\right| + \left|M_{f_i}\right| \tag{7}$$

$$\vec{P}_{tip} = P_{tip}[\sin\beta\,\hat{x} + \cos\beta\,\hat{y}] \tag{8}$$

where $\beta = \theta_n - \theta_{tip}$, the difference between the rotation of the fingertip segment and the inclination of the tip relative to the horizontal in the unrotated fingertip segment, and the total moments due to cable tension, pressure and friction are

$$|M_{T_n}| = |T_n[\sin\beta\,\hat{x} + \cos\beta\,\hat{y}] \times \vec{r}_{T_n}| \qquad (9)$$

$$|M_{F_i}| = \left|\sum_{i=1}^{n} F_i(\sin\theta_i\,\hat{x} + \cos\theta_i\,\hat{y}) \times \vec{r}_{hole,i}\right| \qquad (10)$$

$$|M_{f_i}| = \left|\sum_{i=1}^{n} f_i(\cos\varphi_i\,\hat{x} + \sin\varphi_i\,\hat{y}) \times \vec{r}_{hole,i}\right| \qquad (11)$$

Where, again, all $F_i$, $f_i$ and $T_n$ may be related to the input tension $T_o$ using (4), (3) and (2), respectively.

Lastly, finite element models of the segment were developed to perform stress analysis of the segments in the test the performance during normal use. The static loading simulations of the segments were performed using the SolidWorks 2016 Simulation FEA package, as shown in Fig. 4. Boundary conditions are described in Fig. 4A. Loads imposed by the cable were calculated based on the case of a 5-segment-long phalange forming a 180º circular arc to grasp a cylindrical object, with an assumed cable tension of 44 N, based on force transmission tests described previously.

More comprehensive FEM studies at different segment orientations will be necessary to draw definitive conclusions about the strength of the segments. However, the current round of studies predicted von Mises stresses of 29 MPa around the cable passage, as shown in Fig. 4B. This result was within the average yield strength of the segment material (PLA plastic; 44 MPa), however only by a small factor of safety [13]. While PLA is acceptable for prototyping purposes, a stronger material will be necessary for any commercial versions of the exoskeleton. It may also be advisable to make the cross-bar thicker to increase its strength and stiffness.

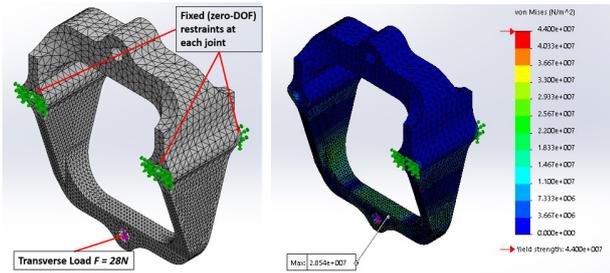

**Fig. 4:** (A) Finite element model of the segment under transverse loading from the flexor cable. (B) FEM results for the segment under transverse loading.

In the force tracking experiment with four subjects shown in Fig. 5 the average forces produced by the wearable robot in the thumb, index fingers, and middle fingers are 31.2 N, 17.8 N, and 21.4 N respectively. It demonstrated that the fingertip force of soft robotic glove produces more than 70% of the torque in healthy biological fingers [14] to assist activities of daily living while ensuring inherent safety without exerting excessive forces to the wearers. This robotic glove produces the greater forces comparing results reported in the literature [4,5,15] to the best of our knowledge.

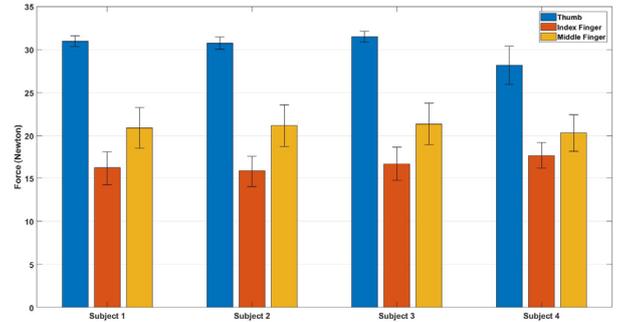

**Fig. 5:** The fingertip force generated by exoskeleton on the thumb, index finger, and middle fingers in 4 subjects.

## CONCLUSION

The hand exoskeleton design and cable transmission mechanism successfully support finger flexion and is capable of achieving fingertip forces of at least 22 N. It is lightweight and contours to the hands, making wearers comfortable. Now the prototype of wearable hand exoskeleton is mainly made by 3d printing materials, the rapid prototype with low layer resolution brings extra friction. Future work on this design will include developing mechanisms to control flexion and extension with a single electric motor per phalange, design a portable mount for the actuator unit, design finger segments with higher traction for gripping, optimize wearable structure quick donning and doffing, and further improve the comfort of the device by designing secure restraints.

## ACKNOWLEDGMENTS

This work is supported by the Grove School of Engineering, The City University of New York, City College. Any opinions, findings, and conclusions or recommendations expressed in this material are those of the author(s) and do not necessarily reflect the views of the funding organization.